\documentclass{sig-alternate}

\toappear{}

\usepackage[overlay,absolute]{textpos}

\begin{document}

\begin{textblock*}{10in}(16mm, 10mm)
{\textbf{Ref:} \emph{ACM Genetic and Evolutionary Computation Conference (GECCO)}, pages 1469--1475, Atlanta, GA, July 2008.}
\end{textblock*}

\begin{textblock*}{10in}(62mm, 16mm)
{\fbox{\large Winner of Best Paper Award in GECCO 2008}}
\end{textblock*}

\conferenceinfo{GECCO'08,} {July 12--16, 2008, Atlanta, Georgia, USA.} 

\CopyrightYear{2008}

\crdata{978-1-60558-131-6/08/07}

\title{Genetic Algorithms for Mentor-Assisted \\Evaluation Function Optimization}

\numberofauthors{3}

\author{
\alignauthor
Eli (Omid) David\titlenote{www.elidavid.com}\\
       \affaddr{Dept.~of Computer Science}\\
       \affaddr{Bar-Ilan University}\\
       \affaddr{Ramat-Gan 52900, Israel}\\
       \email{mail@elidavid.com}\\ 
\alignauthor
Moshe Koppel\\
       \affaddr{Dept.~of Computer Science}\\
       \affaddr{Bar-Ilan University}\\
       \affaddr{Ramat-Gan 52900, Israel}\\
       \email{koppel@cs.biu.ac.il}\\ 
\alignauthor
Nathan S.~Netanyahu\titlenote{Nathan Netanyahu is also affiliated with the Center for Automation Research, University of Maryland, College Park, MD 20742 (e-mail: nathan@cfar.umd.edu).}\\
       \affaddr{Dept.~of Computer Science}\\
       \affaddr{Bar-Ilan University}\\
       \affaddr{Ramat-Gan 52900, Israel}\\
       \email{nathan@cs.biu.ac.il}\\
}

\maketitle

\begin{abstract}
In this paper we demonstrate how genetic algorithms can be used to reverse engineer an evaluation function's parameters for computer chess. Our results show that using an appropriate \emph{mentor}, we can evolve a program that is on par with top tournament-playing chess programs, outperforming a two-time World Computer Chess Champion. This performance gain is achieved by evolving a program with a smaller number of parameters in its evaluation function to mimic the behavior of a superior mentor which uses a more extensive evaluation function. In principle, our mentor-assisted approach could be used in a wide range of problems for which appropriate mentors are available. 
\end{abstract}




\section{Introduction}

Since the dawn of modern computer science, game playing has posed a formidable challenge in the field of Artificial Intelligence. Many founding figures of computer science and AI (including Alan Turing, Claude Shannon, Konrad Zuse, Arthur Samuel, John McCarthy, Ken Thompson, Herbert Simon, and others) developed game-playing programs and used games in AI research.

The ongoing key role played by and the impact of computer games on AI should not be underestimated. If nothing else, computer games have served as an important testbed for spawning various innovative AI techniques in domains and applications such as search, automated theorem proving, planning, and learning. In addition, the annual World Computer Chess Championship (WCCC) is arguably the longest ongoing performance evaluation of programs in computer science, which has inspired other well-known competitions in robotics, planning, and natural language understanding.

Computer chess, while being one of the most researched fields within AI, has not lent itself to the successful application of conventional learning methods, due to its enormous complexity. Hence, current top chess programs use manually tuned parameters for their evaluation function, which is the most critical component of any chess program.

\textbf{In this paper, we introduce a novel \emph{mentor-assisted} approach for automatically evolving the parameters of a chess program's evaluation function through the use of genetic algorithms (GA). The results show that our mentor-assisted approach for application of GA efficiently evolves the parameters of the evaluation function from randomly initialized values to highly tuned ones, yielding a program that outperforms its original version by a wide margin. Such performance was achieved for an evolved program whose evaluation function is considerably more compact than that of the mentor, in terms of its number of parameters.}

In Section 2 we review past attempts at applying evolutionary techniques in computer chess, and also compare alternative learning methods to evolutionary methods, arguing why the latter are more appropriate for the task in question. Section 3 provides our mentor-assisted approach, including a detailed description of the chess programs we used, and the framework of the GA as applied to the problem. Section 4 provides our experimental results, and Section 5 contains concluding remarks and suggestions for future research.

\newpage
\section{Learning in Computer Chess}

The enormously complex game of chess, referred to as ``the touchstone of the intellect'' by Goethe, has always been one of the main battlegrounds of man versus machine (John McCarthy \cite{mccarthy90} refers to chess as the ``Drosophila of AI''). Chess-playing programs have come a long way over the past several decades. While the first chess programs could not pose a challenge to even a novice player, the current advanced chess programs are on par with the strongest human chess players, as the recent man vs.~machine matches clearly indicate. This improvement is largely a result of deep searches possible nowadays, thanks to both hardware speed and improved search techniques. While the search depth of early chess programs was limited to only a few plies, nowadays tournament-playing programs easily search more than a dozen plies in middlegame, and tens of plies in late endgame.

Despite their groundbreaking achievements, a glaring deficiency of all top chess programs today is their near total lack of a learning capability (except in most negligible ways, e.g., ``learning'' not to play an opening that resulted in a loss, etc.). The resulting criticism is that despite their intelligent behavior, these top chess programs have no underlying intelligence, and are mere brute-force (albeit efficient) searchers.

\subsection{Conventional vs. Evolutionary Learning in Computer Chess}

During more than fifty years of research in the area of computer games, many learning methods such as reinforcement learning \cite{sutton98} have been employed in simpler games. Temporal difference learning has been successfully applied in backgammon and checkers \cite{schaeffer01,tesauro92}. Although temporal difference learning has also been applied to chess \cite{baxter00}, the results showed that after three days of learning, the playing strength of the program was merely 2150 Elo (see Appendix B for a description of the Elo rating system), which is a very low rating for a chess program. Wiering \cite{wiering95} provided formal arguments for the failure of these methods in more complicated games such as chess.

The problem of learning in computer chess is basically an optimization problem. Each program plays by conducting a search, with the current position being the root of the search tree, and applying static evaluation function at leaf nodes. In other words, sophisticated as the search algorithms may be, most of the knowledge of the program is in its evaluation function. Even though automatic tuning methods, based mostly on reinforcement learning, have been successfully applied to other simpler games such as checkers, none of them have found their way into the state-of-the-art chess engines. Currently, all top tournament-playing chess programs use hand tuned evaluation functions, since conventional learning methods cannot cope with the enormous complexity.

We believe that genetic algorithms are best suited for automatic tuning of an evaluation function's parameters, for the following reasons:

(1) The space to be searched is huge. It is estimated that there are up to $10^{200}$ possible positions that can arise in chess. As a result, any method based on exhaustive search of the problem space is infeasible.

(2) The search space is not smooth and unimodal. An evaluation function's parameters in any top chess program are highly co-dependent. For example, in many cases increasing the values of three parameters will result in a worse performance, but if a fourth parameter is also increased, then an improved overall performance would be obtained. Since the search space is not unimodal, i.e., does not consist of a single smooth ``hill'', any gradient-ascent algorithm such as hill climbing will perform poorly.

(3) The problem is not well understood. As will be discussed in detail in the next section, even though all top performing programs are hand tuned by their programmers, finding the best value for each parameter is mostly based on educated guessing and intuition. (The fact that all top programs continue to operate in this manner, attests to the lack of practical alternatives.) Had the problem been well understood, a domain-specific heuristic would have outperformed a general-purpose method such as GA.

(4) We do not require a global optimum to be found. Our goal in tuning an evaluation function is to adjust its parameters so that the overall performance of the program is enhanced. In fact, a unique global optimum does not exist for this tuning problem.

At first glance, automatic tuning of the evaluation function appears like an optimization task, well suited for GA. The many parameters of the evaluation function (bonuses and penalties for each property of the position) can be encoded as a bit-string. We can randomly initialize many such ``chromosomes'', each representing one evaluation function. Thereafter, one needs to evolve the population until highly tuned ``fit'' evaluation functions emerge.

However, there is one major obstacle that hinders the application of GA, namely the fitness function. Given a set of evaluation parameters (encoded as a chromosome), how should the fitness value be calculated? For many years, it seemed that the solution was to let the individuals in each generation (each individual being a chess program with the appropriate evaluation function) play against each other a series of games, and subsequently, take the score of each individual as its fitness value.

The main drawback of this approach is the unacceptably large amount of time needed to evolve each generation. As a result, severe limitations are imposed on the length of the games played after each generation, and also on the size of the population involved. With a population size of 100, and by an extreme limitation of 1 minute per game for each side, assuming each individual plays at least 10 games, it would take 2000 minutes for each generation to evolve. With these time constraints, reaching the 50th generation would take no less than 70 days.

In the next section we present our mentor-assisted approach for using GA in state-of-the-art chess programs. Before that, we briefly review previous work in applying evolutionary methods in computer chess.

\subsection{Previous Evolutionary Methods Applied \\to Chess}

Despite the abovementioned problems, there have been some successful applications of evolutionary techniques in computer chess, subject to some restrictions. Genetic programming was successfully employed by Hauptman and Sipper \cite{hauptman05,hauptman07} for evolving programs that can solve Mate-in-N problems and play chess endgames.

Kendall and Whitwell \cite{kendall01} used evolutionary algorithms for tuning evaluation function parameters. Their approach had limited success, due to the very large number of games required (as previously discussed), and the small number of parameters used in their evaluation function. Their evolved program managed to compete with strong programs only if their search depth (lookahead) was severely limited.

Similarly, Aksenov \cite{aksenov04} used genetic algorithms for evolving evaluation function parameters, using games between the organisms for determining their fitness. Again, since this method required a very large amount of games, the method evolved only a few evaluation function parameters with limited success. Tunstall-Pedoe \cite{tunstall91} also suggested a similar approach, without providing an implementation.

Gross \emph{et al.}~\cite{gross02} used a hybrid of genetic programming and evolution strategies to improve the efficiency of an already existing search algorithm using a distributed computing environment on the internet.

In the following section, we present a novel approach that facilitates the use of GA for efficiently evolving an evaluation function's parameters. As our results will demonstrate, this method is very fast, and the evolved program is on par with the world's strongest chess programs. 

\section{Mentor-Assisted Fitness \\Evaluation}

Due to the impediments already discussed, establishing fitness evaluation by means of playing numerous games is not practical. However, we can exploit a vast reservoir of previously under-utilized information. While the evaluation functions of existing chess programs are carefully guarded secrets, it is standard practice for a chess program to (partially) reveal the score for any given position encountered in a game. We propose to use genetic algorithms to essentially reverse engineer these evaluation functions. We will see that such reverse engineering can be carried out very rapidly and successfully, and that in fact a program based on an evaluation function learned from a particular mentor, can perform as well as the mentor. Specifically, the program evolves its evaluation function from a mentor using the steps appearing in Figure \ref{fig:mentor}.

\begin{figure}[htbp]
\begin{center}
\line(1,0){240}

\begin{enumerate}

\item Generate a list of random problems.
\item For each problem, let the mentor evaluate the problem and store the result.
\item Let each individual evaluate all the problems, and for each problem calculate the average difference between the value given by the individual and the value issued by the mentor. The fitness of the individual will be inversely proportional to this average difference.

\end{enumerate}

\line(1,0){240}
\caption{Mentor-assisted fitness evaluation.}
\label{fig:mentor}
\end{center}
\end{figure}

In our case, each problem is associated with a chess position, and the mentor is the evaluation function of a state-of-the-art chess engine. In other words, we generate a list of random chess positions for each generation, and let a strong chess engine evaluate all of them. Afterwards, we let the evaluation function of each of these individuals evaluate the positions. The closer the evaluation of an individual to the evaluation of the mentor is, the higher its fitness value will be. In the following subsections, we describe in detail the chess programs, the implementation of our mentor-assisted approach, and the GA parameters used.

\subsection{The Chess Programs}

For this work, we used the \textsc{Maestro} and \textsc{Falcon} chess programs. \textsc{Falcon} is a 2700+ Elo rated grandmaster-level chess program, which has successfully participated in two World Computer Chess Championships. (See Appendix B for Elo rating system.) \textsc{Falcon} uses \textsc{NegaScout}/PVS \cite{campbell83,reinfeld83} search, with null-move pruning \cite{beal89,david02,donninger93}, internal iterative deepening \cite{anantharaman91,scott69}, dynamic move ordering (history + killer heuristic) \cite{akl77,gillogly72,schaeffer83,schaeffer89}, multi-cut pruning \cite{bjornsson98, bjornsson01}, selective extensions \cite{anantharaman91,beal95} (consisting of check, one-reply, mate-threat, recapture, and passed pawn extensions), transposition table \cite{nelson85,slate77}, futility pruning near leaf nodes \cite{heinz98a}, and blockage detection in endgames \cite{david06}. 

\textsc{Falcon} has an extensive evaluation function consisting of several thousand lines of code, using more than 100 evaluation function parameters. \textsc{Maestro}, an experimental program which is considerably weaker than \textsc{Falcon}, employs identical search techniques, and differs from \textsc{Falcon} only by its evaluation function. \textsc{Maestro}'s evaluation function is much smaller, consisting of fewer than 40 parameters. Since the two programs differ only in their evaluation function, this element is solely responsible for their vastly different performances.

\subsection{Encoding the Evaluation Function}

We use the stronger program, \textsc{Falcon}, as the mentor, and evolve the evaluation function of \textsc{Maestro} to mimic the behavior of its mentor, thereby improving its strength. We use only the output of \textsc{Falcon}'s evaluation function, and otherwise assume we know nothing about the methods \textsc{Falcon} uses to compute this function. Thus, we use \textsc{Falcon}'s scores to optimize the parameters of \textsc{Maestro}, not the parameters of \textsc{Falcon}, which (for our purposes) are unknown to us.

While having a simpler evaluation function than \textsc{Falcon}'s, \textsc{Maestro}'s evaluation function does cover all important aspects of a position. The main features of an evaluation function are material, piece mobility and centricity, pawn structure, and king safety. We show that comparable performance can be achieved with a considerably smaller number of parameters than that of the mentor's evaluation function.

In order to demonstrate the effectiveness of this mentor-based approach, we chose to entirely ignore the initial values of the evaluation function parameters, and instead, assign random values to all of them. In other words, if \textsc{Maestro} was 120 Elo weaker than \textsc{Falcon}, after initializing randomly its evaluation function parameters, it will play like a novice that has no knowledge about the game (apart from making legal moves and certain built-in tactics).

We represent the evaluation function parameters of \textsc{Maestro} as a binary bit-string (chromosome size: 230 bits), initialized randomly. In addition to the random initialization, we further impose the following restriction: Except for the five parameters representing the material values of the pieces, all the other parameters were assigned a fixed length of 6 bits per parameter. Obviously, there are many parameters for which 3 or 4 bits will suffice. However, we allocated a fixed length of 6 bits to all parameters so that \emph{a priori} knowledge would not bias the algorithm in any way.

\subsection{Mentor-Assisted Fitness Function}

As already described, our goal is to evolve the parameters of \textsc{Maestro} so that its evaluation function would produce as close a score as possible to the evaluation function of \textsc{Falcon}, given the same position. 

For our experiments, we used a database of 10,000 grandmaster games (rating above 2600 Elo), and randomly chose one position from each game. Of these 10,000 positions, we selected 5,000 positions for training, and 5,000 for testing. For all of the positions we assumed it was White's turn to move (since this is more likely to permit materially imbalanced positions in our sample).

At first, we let \textsc{Falcon} search each of the 10,000 positions to a depth of 2 plies, and stored its evaluation of the position. (We denote this mentor's score for position $p$ by $S_{m,p}$.) Then, in each generation we randomly selected 1,000 positions out of the 5,000 designated positions for the learning phase. This random selection of positions in each generation introduces additional variety, which should help prevent premature convergence to suboptimal values.

For each organism we translate its chromosome bit-string into a corresponding evaluation function, and apply the evaluation function to each of $N$ positions examined (in our case $N=1000$). Organism $i$'s score for position $p$ is denoted by $S_{i,p}$. For each position $p$ we define the organism's error as

\begin{displaymath}
E_{i,p} = |S_{m,p} - S_{i,p}|,
\end{displaymath}
 
and the overall error for the organism, $E_i$, will be the average error over the $N$ positions, i.e.,

\begin{displaymath}
E_i = \frac{\displaystyle \sum_{p=1}^{N} E_{i,p}}{N}.
\end{displaymath}

Finally, the fitness value of organism $i$ will be $F_i = -E_i$, i.e., the smaller the average error, the higher the fitness value.

\subsection{GA Parameters}

Apart from the special fitness function described above, we used a standard implementation of GA with proportional selection and single point crossover. The following are the parameters we used:
\\
\\
population size = 1000\\
crossover rate = 0.75\\
mutation rate = 0.002\\
number of generations = 300\\

At the end of each generation, we replicated the best organism, and deleted the worst organism.

Note that each organism is in fact a unique encoding of the evaluation function values, which is loaded by the \textsc{Maestro} engine.

In the following section we provide our experimental results, both in terms of the learning efficiency, and the performance gain of the best evolved individual.

\section{Results}

We first present the results of running the mentor-assisted GA as described in the previous section. Then, we provide the results of several experiments that measure the strength of the evolved program in comparison to its original version.

\subsection{Learning Results}

\begin{figure}[htbp]
\centering
\epsfig{file=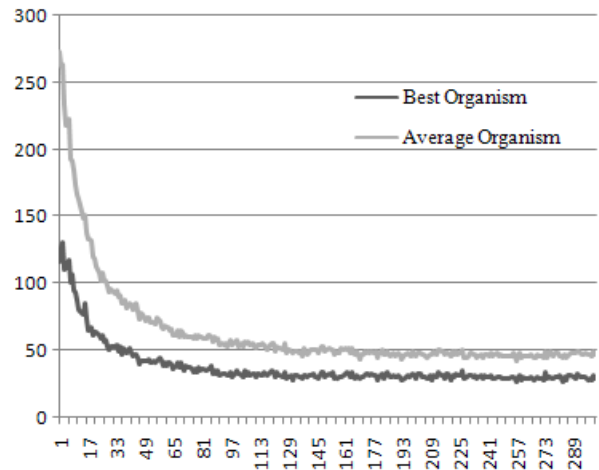, height=2.5in, width=3.1in}
\caption{Average error per position for the best organism and the population average in each generation (total time for 300 generations: 442 seconds).}
\label{fig:graph}
\end{figure}

Figure~\ref{fig:graph} shows the average error per position of the best organism and the population average for 300 generations. An evaluation unit in chess programs is commonly called a \emph{centipawn}, i.e., 1/100th of the value of a pawn. Traditionally, a pawn value is assigned a value of 100, and all the other figures are assigned relative values to that of the pawn. However, the value of pawn itself need not necessarily be 100, so a unit of evaluation may no longer necessarily be equal to 1/100th of a pawn. Despite this inconsistency, the term centipawn is still used to denote the smallest evaluation unit.

The results in Figure~\ref{fig:graph} show that in the first few generations the average error was more than 250 centipawns, and the best organism's error was more than 130. This huge initial error is not surprising, as we initialized all the values randomly (i.e., the initial individual did not even know that a queen is worth more than a pawn). This lack of initial knowledge results in very large initial errors, which correspond to very low fitness values for the organisms. This results in the rapid extinction of individuals with highly erroneous parameter values. As early as generation 35, the best organism's average error drops below 50 centipawns. At this stage, large parameter values (such as material values, etc.) are already well-tuned for most of the organisms, and for the remaining generations, the smaller values are fine tuned. At generation 300, the average error of the best organism is 28 centipawns, and the average error of the population is 47 centipawns. Figure~\ref{fig:best-evolved} provides the evolved values of the best individual.

\begin{figure}[ht!]
\begin{center}
\line(1,0){240}
\begin{verbatim}
     PAWN_VALUE                          83
     KNIGHT_VALUE                       322
     BISHOP_VALUE                       323
     ROOK_VALUE                         478
     QUEEN_VALUE                        954
     PAWN_ADVANCE_A                       2
     PAWN_ADVANCE_B                       4
     PASSED_PAWN_MULT                     5
     DOUBLED_PAWN_PENALTY                21
     ISOLATED_PAWN_PENALTY               10
     BACKWARD_PAWN_PENALTY                3
     WEAK_SQUARE_PENALTY                  7
     PASSED_PAWN_ENEMY_KING_DIST          5
     KNIGHT_SQ_MULT                       7
     KNIGHT_OUTPOST_MULT                  8
     BISHOP_MOBILITY                      5
     BISHOP_PAIR                         44
     ROOK_ATTACK_KING_FILE               30
     ROOK_ATTACK_KING_ADJ_FILE            1
     ROOK_ATTACK_KING_ADJ_FILE_ABGH      21
     ROOK_7TH_RANK                       32
     ROOK_CONNECTED                       2
     ROOK_MOBILITY                        2
     ROOK_BEHIND_PASSED_PAWN             48
     ROOK_OPEN_FILE                      12
     ROOK_SEMI_OPEN_FILE                  6
     ROOK_ATCK_WEAK_PAWN_OPEN_COLUMN      7
     ROOK_COLUMN_MULT                     3
     QUEEN_MOBILITY                       0
     KING_NO_FRIENDLY_PAWN               27
     KING_NO_FRIENDLY_PAWN_ADJ           17
     KING_FRIENDLY_PAWN_ADVANCED1        12
     KING_NO_ENEMY_PAWN                  11
     KING_NO_ENEMY_PAWN_ADJ               3
     KING_PRESSURE_MULT                   8
\end{verbatim}

\line(1,0){240}
\caption{Evolved parameters of the best individual.}
\label{fig:best-evolved}
\end{center}
\end{figure}

With the completion of the learning phase, we used the additional 5,000 positions set aside for testing. We let our best evolved organism evaluate these positions, and compared its evaluation with that of its mentor (\textsc{Falcon}). The average error here was 30 centipawns. This indicates that the first 5,000 positions we used for training cover most types of positions that can arise, as the average error is very similar to the average error for the testing set. The entire 300 generation evolution lasted 442 seconds on our machine (see Appendix A), that is, less than 8 minutes.

The results clearly demonstrate that the evolution converges very quickly to values that are very close to those of the mentor. In other words, within a few minutes we evolved an evaluation function from random values to values that result in a behavior closely resembling that of the mentor.

\subsection{Performance of the Evolved Organism}

In this subsection we provide the results of a series of matches between the programs. All matches consisted of 300 games at a time control of 3 minutes per game for each side. We are interested in comparing the performance of the original \textsc{Maestro} vs.~\textsc{Falcon} (in order to obtain a baseline), and the performance of the evolved program (which we will call \textsc{MaestroGA}) against its mentor, \textsc{Falcon}. We also compare the evolved \textsc{MaestroGA} against the original \textsc{Maestro}. Table~\ref{tab:results} provides the results of the 900 games played. 

\begin{table}[htbp]

\begin{center}
\begin{tabular}{|l||c|c|c|}
\hline
Match & Result & W\% & RD\\
\hline
\hline
\textsc{Falcon} - \textsc{Maestro} & 199.0 - 101.0 & 66.3\% & +118\\
\hline
\textsc{Falcon} - \textsc{MaestroGA} & 151.5 - 148.5 & 50.5\% & +3\\
\hline
\textsc{MaestroGA}-\textsc{Maestro} & 202.5 - 97.5 & 67.5\% & +127\\
\hline
\end{tabular}
\end{center}
\vspace*{-10pt} \caption{Results of 900 games between the three programs (W\% is the winning percentage, and RD is the Elo rating difference). Win = 1 point, draw = 0.5 point, and loss = 0 point.}
\label{tab:results}

\end{table}

The results of the match between \textsc{Falcon} and \textsc{Maestro} show that \textsc{Falcon} is considerably stronger than \textsc{Maestro}. As we mentioned earlier, these two programs differ only in their evaluation function, with \textsc{Falcon} having a larger evaluation function.

It should be noted that in terms of its choice of parameters and in all respects other than the values assigned to each parameter, \textsc{MaestroGA} is basically the original \textsc{Maestro} program\footnote{Both \textsc{Maestro} and \textsc{MaestroGA} are in fact the very same executable file, with the only difference being that \textsc{MaestroGA} loads the values of its evaluation function's parameters from an external ``chromosome file''.}. Despite having a smaller evaluation function, our mentor-assisted GA evolves parameter values such that \textsc{MaestroGA}'s performance is essentially identical to that of the mentor, \textsc{Falcon}. In head to head competition against \textsc{Falcon}, it proves to be \textsc{Falcon}'s equal. Moreover, the evolved \textsc{MaestroGA} resoundingly defeats the original version by scoring 67.5\%.

Our results clearly demonstrate the importance of efficient automatic parameter tuning. Despite the fact that the evaluation function parameters in the original \textsc{Maestro} were manually tuned, automatic tuning that started from random values managed to produce much better results than those do human-picked values, resulting in a vast rating difference. 

Finally, we ran three additional matches each consisting of 300 games against the chess program \textsc{Crafty}, by Robert Hyatt \cite{hyatt90}. \textsc{Crafty} has successfully participated in numerous World Computer Chess Championships (WCCC), and is a direct descendent of \textsc{Cray Blitz}, the winner of 1983 and 1986 WCCC. \textsc{Crafty} is frequently used in the literature as a standard reference. Thus, we compared our original \textsc{Maestro}, the evolved \textsc{MaestroGA}, and the mentor, \textsc{Falcon}, against \textsc{Crafty}. Table~\ref{tab:crafty} provides the results.

\begin{table}[htbp]

\begin{center}
\begin{tabular}{|l||c|c|c|}
\hline
Match & Result & W\% & RD\\
\hline
\hline
\textsc{Maestro} - \textsc{Crafty} & 140.5 - 159.5 & 46.8\% & $-22$\\
\hline
\textsc{MaestroGA} - \textsc{Crafty} & 177.0 - 123.0 & 59.0\% & +63\\
\hline
\textsc{Falcon} - \textsc{Crafty} & 173.5 - 126.5 & 57.8\% & +55\\
\hline
\end{tabular}
\end{center}
\vspace*{-10pt} \caption{CRAFTY vs.~MAESTRO, MAESTROGA, and FALCON (W\% is the winning percentage, and RD is the Elo rating difference).}
\label{tab:crafty}

\end{table}

The results show that while \textsc{Crafty} is stronger than the original \textsc{Maestro}, the evolved version is clearly superior to \textsc{Crafty}. Interestingly, \textsc{MaestroGA}'s performance against \textsc{Crafty} is actually marginally better than that of its mentor.

In another experiment, we again found that in certain ways the evolved program can marginally outperform the mentor. For measuring the tactical strength of the programs, we used the Encyclopedia of Chess Middlegames (ECM) test suite, consisting of 879 positions. Each program was given 5 seconds per position. Table~\ref{tab:ecm} provides the results. As can be seen, \textsc{MaestroGA} solved significantly more problems than did \textsc{Maestro} and even a few more than \textsc{Falcon}. 

\begin{table}[htbp]
\begin{center}
\begin{tabular}{|c|c|c|}
\hline
\textsc{Maestro} & \textsc{MaestroGA} & \textsc{Falcon}\\
\hline
\hline
616 & 649 & 645\\
\hline
\end{tabular}
\end{center}
\vspace*{-10pt} \caption{Number of ECM positions solved by each program (time: 5s per position).}
\label{tab:ecm}
\end{table}

In summary, the results show that our mentor-assisted approach for application of GA allows for an efficient evolution of the parameters to exhibit a behavior that is very similar to that of the mentor. Even though we evolved the parameters from scratch, with the chromosome being initialized randomly, the evolved values proved to perform substantially better than the hand tuned values in the original version. The results of the 1800 games played, demonstrate the marked performance superiority of the evolved version over the original version.

\section{Concluding Remarks and\\ Future Research}

In this paper we presented a novel mentor-assisted approach for automatic tuning of parameters. Wherever an already intelligent entity exists, it can serve as a mentor, such that the GA will evolve the organisms to mimic the behavior of the mentor. In other words, our approach enables the duplication of another intelligent organism's behavior, merely by looking at its behavior, with no access to the underlying mechanism of this mentor.

Our experimental results showed that within a few minutes organisms were evolved from a randomly initialized chromosome, to highly tuned parameters that produced similar behavior to that of the mentor, in terms of the relative performance observed with respect to the same set of positions. The results of the games played demonstrated the significant gain of the evolved version, which clearly outperformed its original version. Note that the successful duplication of the mentor's behavior was achieved despite the fact that a considerably smaller number of parameters were used in the evaluation function of the evolved program.

For future research, we intend to develop further capabilities based on the presented mentor-assisted approach. In this paper we showed how another computer program can serve as a mentor. However, using human players as mentors is more difficult, as in this case we do not have access to their numerical evaluation of the position. We believe, though, that a database containing hundreds of games of a human player will provide sufficient data for this learning to take place. One method we intend to experiment with, is to extract several thousand positions from games played by a human mentor, and for each position assign a higher fitness for the organism that produces the move played by the mentor. This approach, if successful, would basically enable the program to behave like its mentor, without having access to his/her ``brain''. For example, we might be able to develop a program that plays like Kasparov by learning from his games.

In this work we used a single mentor. An alternative implementation may use several mentors, using the ``wisdom of crowds'' concept to evolve an individual which is ``wiser'' than its mentors. It is well-known that each chess program has its strengths and weaknesses. By using several mentor chess engines, it might be possible to combine the strengths of all of them, and outperform each individual mentor.

Our mentor-assisted approach could also be applied to the problem of player recognition. Given a list of $N$ players, the simplest approach is to separately evolve $N$ organisms, each mimicking the behavior of the respective player. Then, given a game played by one of the players, each of our $N$ organisms would apply their evaluation to the position, and the player whose cloned organism agrees more closely with the moves played in the game, is more likely to be the player in the game.

Finally, we believe the method of GA-based parameter tuning suggested here could be applied to a wide array of problems in which the output of a mentor's evaluation function is available for training purposes.

\appendix

\section{Experimental Setup}

\hspace*{-4pt}Our experimental setup consisted of the following resources:

\begin{itemize}
\item \textsc{Maestro} and \textsc{Falcon} chess engines running under UCI protocol, and \textsc{Crafty 19.19} running as a native ChessBase engine.

\item Encyclopedia of Chess Middlegames (ECM) test suite, consisting of 879 positions.

\item Fritz 8 interface for automatic running of matches. Fritz opening book was used for all games.

\item AMD Athlon 64 3200+ with 1 GB RAM and Windows XP operating system.
\end{itemize}

\section{Elo Rating System}

The Elo rating system, developed by Prof.~Arpad Elo, is the official system for calculating the relative skill levels of players in chess. The following statistics from the January 2008 FIDE rating list provide a general impression of the meaning of Elo ratings:

\begin{itemize}

\item 18305 players have a rating above 2200 Elo.

\item 2037 players have a rating between 2400 and 2499, most of whom have either the International Master (IM) or the Grandmaster (GM) title.

\item 577 players have a rating between 2500 and 2599, most of whom have the GM title.

\item 141 players have a rating between 2600 and 2699, all but one of whom have the GM title.

\item 24 players have a rating between 2700 and 2799.
\end{itemize}

Only four players have ever had a rating of 2800 or above. A novice player is generally associated with rating values of below 1400 Elo. Given the rating difference ($RD$) of two players, the following formula calculates the expected winning rate ($W$, between 0 and 1) of the player:

\begin{displaymath}
W = \frac{1}{10^{-RD/400} + 1}
\end{displaymath}

Given the winning rate of a player, as is the case in our experiments, the expected rating difference can be derived from the above formula:

\begin{displaymath}
RD = -400 \log_{10}(\frac{1}{W} - 1)
\end{displaymath}

\end{document}